\documentclass[lettersize,journal]{IEEEtran}

\usepackage[caption=false,font=normalsize,labelfont=sf,textfont=sf]{subfig}
\usepackage{amsmath,amssymb,amsfonts}
\usepackage[ruled,vlined]{algorithm2e}
\usepackage{array}
\usepackage{longtable}
\usepackage{textcomp}
\usepackage{booktabs,multirow}
\usepackage{stfloats}
\usepackage{url}
\usepackage{verbatim}
\usepackage{graphicx}
\usepackage{cite}
\usepackage{pifont}
\usepackage{float}
\usepackage{amsthm}

\begin{document}

\title{Online Spatio-Temporal Correlation-Based Federated Learning for Traffic Flow Forecasting}

\author{Qingxiang Liu,
        Sheng Sun,
        Min Liu, ~\IEEEmembership{Senior Member,~IEEE,}
        Yuwei Wang,
        and Bo Gao
\thanks{This work was supported by the National Key Research and Development Program of China (2021YFB2900102) and the National Natural Science Foundation of China (No. 62072436 and No.62202449).}

\thanks{Qingxiang Liu, Sheng Sun, Min Liu and Yuwei Wang are with the Institute of Computing Technology, Chinese Academy of Sciences, Beijing, China. Qingxiang Liu is also with University of Chinese Academy of Sciences, Beijing, China. Min Liu is also with the Zhongguancun Laboratory (e-mail: qingxiangliu737@gmail.com, sunsheng@ict.ac.cn, liumin@ict.ac.cn and ywwang@ict.ac.cn).}

\thanks{Bo Gao is with the School of Computer and Information Technology, Beijing Jiaotong University, Beijing, China (email: bogao@bjtu.edu.cn).}

\thanks{Min Liu is the corresponding author.}
        
}

\markboth{
IEEE Transactions on Intelligent Transportation Systems
}%
{Liu \MakeLowercase{\textit{et al.}}: Online Spatio-Temporal Correlation-Based Federated Learning for Traffic Flow Forecasting}


\maketitle

\begin{abstract}
Traffic flow forecasting (TFF) is of great importance to the construction of Intelligent Transportation Systems (ITS). To mitigate communication burden and tackle with the problem of privacy leakage aroused by centralized forecasting methods, Federated Learning (FL) has been applied to TFF. However, existing FL-based approaches employ batch learning manner, which makes the pre-trained models inapplicable to subsequent traffic data, thus exhibiting subpar prediction performance. In this paper, we perform the first study of forecasting traffic flow adopting Online Learning (OL) manner in FL framework and then propose a novel prediction method named Online Spatio-Temporal Correlation-based Federated Learning (FedOSTC), aiming to guarantee performance gains regardless of traffic fluctuation. Specifically, clients employ Gated Recurrent Unit (GRU)-based encoders to obtain the internal temporal patterns inside traffic data sequences. Then, the central server evaluates spatial correlation among clients via Graph Attention Network (GAT), catering to the dynamic changes of spatial closeness caused by traffic fluctuation. Furthermore, to improve the generalization of the global model for upcoming traffic data, a period-aware aggregation mechanism is proposed to aggregate the local models which are optimized using Online Gradient Descent (OGD) algorithm at clients. We perform comprehensive experiments on two real-world datasets to validate the efficiency and effectiveness of our proposed method and the numerical results demonstrate the superiority of FedOSTC.
\end{abstract}


\section{Introduction}
The development of Internet of Things (IoT) has quicken the construction progress of Intelligent Transport Systems (ITS), which has aroused great interests in industry and academia\cite{lin2020spatiotemporal,zhu2019parallel}. In recent years, an increasing number of traffic nodes, e.g., traffic sensors and loop detectors are mounted along roads to generate overwhelming data for enhanced traffic service, e.g., traffic flow forecasting \cite{liu2020dynamic,chen2020edge}.
By far, plenty of deep learning-based methods are proposed to improve performance gains in traffic flow forecasting \cite{lv2014traffic,tan2019recognizing}.

Recurrent Neural Network (RNN) and its variants, e.g., Long Short-Term Memory network (LSTM) \cite{zhao2017lstm} and Gated Recurrent Unit network (GRU) \cite{li2017diffusion,zhang2018combining}, are widespreadly used in forecasting traffic flow, due to the effectiveness of capturing temporal patterns inside traffic data sequences.
Besides, these methods combined with graphing approaches, such as Graph Neural Network (GNN) \cite{li2021hierarchical} or Graph Attention Network (GAT) \cite{wu2018graph}, can extract spatial correlation among traffic nodes, thus further increasing forecasting accuracy.
Despite prominent performance improvement, these methods are based on centralized learning strategy, which involves transmitting large quantities of privately-owned traffic data from decentralized traffic nodes to the central server for training prediction models.
This strategy yields considerable communication overhead and privacy concern, since private information (e.g., plate numbers) contained in traffic data may be leaked during data exchange. 

As a result, it is crucial to forecast traffic flow with keeping traffic data decentralized for mitigating communication burden and preserving privacy.
Federated Learning (FL), as a novel distributed computing paradigm can resolve the problem above, where traffic nodes (called clients) collaboratively train a prediction model based on their private data, and only exchange intermediate model parameters with the server for model aggregation \cite{yang2019federated,li2020federated}.
Owing to the merit of guaranteeing competitive forecasting performance without direct data transmission, there have been a few efforts in forecasting traffic flow depicting spatio-temporal correlation based on FL framework \cite{liu2020privacy,zhang2021fastgnn,meng2021cross}.

However, all of these approaches employ batch learning manner, where prediction models are trained in advance based on batches of historical traffic data. 
This coarse-grained manner of optimizing prediction models makes the optimized models fail to extensively capture the hidden temporal patterns in traffic flows. 
Furthermore, due to traffic fluctuation, these optimized models easily suffer from poor prediction performance, when directly applied to the upcoming traffic data.
Therefore, these pre-trained models need to be repetitively trained, which incurs extra computation consumption, leads to delayed prediction, and fails to fulfill real-time prediction.

Online Learning (OL) manner can break through the above-mentioned limitations, in which traffic nodes consistently update prediction models and preform predictions once they observe new traffic data. 
Naturally, OL manner has the inherent advantages of real-time prediction and non-redundant deployment and hence applies to the task of traffic flow forecasting.
Traffic flows fluctuate severely at some time stamps, which poses great challenges to prediction models updated in batch learning manner.
However, in OL manner, the fine-grained optimization mechanism of prediction models can adapt to the dynamic changes of temporal patterns hidden in traffic flows, thus guaranteeing the prediction performance.
\textbf{To the best of our knowledge, there are no researches focusing on traffic flow forecasting employing OL manner in FL framework} and this article fills this research gap.


However, there remain two main challenges to be tackled. 
First, we should dynamically evaluate the spatial correlation among clients at each round so as to guarantee prediction performance, which remarkably increases the difficulty of prediction and model update. Second, existing FL approaches adopt the averaging mechanism to yield fresh global models which have poor performance on the subsequent traffic data, due to traffic fluctuation. 
Therefore, it is necessary to explore the fluctuation patterns of traffic flow in the process of model aggregation so as to increase the generalization ability of global models and eventually improve performance gains.

To this end, we propose a novel prediction method named \textbf{O}nline \textbf{S}patial-\textbf{T}emporal \textbf{C}orrelation-based \textbf{F}ederated \textbf{L}earning (FedOSTC) for traffic flow forecasting, which possesses excellent capabilities of dynamically capturing spatio-temporal dependence and boosting model adaptation regardless of traffic fluctuation.
Specifically, a GRU with encoder-decoder architecture is regarded as the prediction model and trained collaboratively by multiple traffic nodes.
The temporal patterns inside traffic flows are captured independently via encoders at clients (traffic nodes). 
Then, the central server adaptively evaluates spatial closeness from temporal patterns using GAT, which can mitigate the effect of traffic fluctuation on spatial correlation.
Relying on the instantly-captured spatio-temporal correlation, clients perform prediction and then incrementally optimize model parameters using Online Gradient Descent (OGD).
Finally, based on the observation of periodic patterns of traffic flows, we generate fresh global models using the newly-developed period-aware aggregation mechanism rather than averaging mechanism, which can increase the generalization of global models for subsequent traffic flows.
The main contributions of this paper are described as follows:

\begin{itemize}
	\item{ We propose a novel prediction model named FedOSTC for forecasting traffic flow in FL framework, to tackle with the problems of high communication burden and privacy leakage in centralized methods. Specifically, FedOSTC includes online evaluating spatio-temporal correlation and incrementally optimizing prediction models.}
	\item{ We obtain temporal patterns inside traffic data at clients using GRU-based encoders, and the central server adopts GAT to dynamically evaluate spatial closeness based on temporal patterns transmitted from clients, aiming to adapt to the dynamic changes of spatial correlation aroused by traffic fluctuation.}
	\item{ We incrementally update local models using OGD and further propose a period-aware aggregation mechanism, which can increase the generalization of global models and eventually improve performance gains.}
	\item{We conduct comprehensive experiments to validate that our proposed FedOSTC outperforms the related state-of-the-art methods from the perspectives of prediction performance and model generalization ability.}
\end{itemize}

The remainder of this paper is organized as below. 
Section II reviews the related literature.
The problem of traffic flow forecasting adopting OL manner in FL framework is formulated in Section III. 
Then in Section IV, detailed information of our proposed method FedOSTC is elaborated.
Furthermore, extensive experiments are conducted in Section V. In the end, the paper is concluded in Section VI.

\section{Related Work}

\subsection{Traffic Flow Forecasting}
A great variety of methods have been proposed to increase the accuracy of traffic flow forecasting. 
Overall, existing methods can be divided into two categories: parametric and non-parametric methods.
AutoRegressive Integrated Moving Average (ARIMA) and its variants are classical parametric methods and have been researched extensively \cite{lee1999application, williams2003modeling}. 
However, this series of methods have limitations in regressing traffic flows with high fluctuation, because they are based on the assumption that traffic flows have stationary distribution, which is not the case in reality. 
Non-parametric methods are based on machine learning models, e.g., Graph Network (GN) \cite{wu2020comprehensive} and RNN \cite{ma2020daily}. 
GN-based methods are devoted to evaluating spatial correlation among multiple nodes based on the constructed topology network, while RNN can capture temporal dependence inside traffic sequences. 
Therefore, many researches combine GN approaches and RNN to measure spatio-temporal correlation and have achieved satisfactory performance \cite{guo2019attention, zhu2020novel}.

In \cite{guo2019attention}, Guo \emph{et al.} proposed an Attention based Spatio-Temporal Graph Convolutional Network (ASTGCN), which contains a spatio-temporal attention mechanism for dynamically capturing the spatial patterns and temporal features. 
In \cite{zhu2020novel}, Zhu \emph{et al.} proposed a traffic prediction method based on Graph Convolutional Network (GCN) and RNN, where the topology graph of road network is utilized for depicting the spatial correlation, aiming to increase forecasting accuracy. 

However, these researches all adopt the centralized training strategy, where the central server updates prediction models directly based on traffic data transmitted from decentralized traffic nodes, thus resulting in huge communication overhead and privacy leakage.

\subsection{Federated Learning for Traffic Flow Forecasting}
Some recent researches pay attention to forecasting traffic flow using FL \cite{qi2021privacy,liu2020privacy,zhang2021fastgnn,meng2021cross}. 
In \cite{qi2021privacy}, Qi \emph{et al.} treated the vehicles as clients and introduced the blockchain to avoid the single point of failure. This scenario is not similar to ours.
In \cite{liu2020privacy,zhang2021fastgnn}, Liu \emph{et al.} and Zhang \emph{et al.} investigated a similar scenario where traffic sensors mounted along  roads belong to some certain organizations. 
The traffic data detected by sensors should be directly transmitted to the organizational servers for training, also confronting with excessive communicating burden. 
In order to preserve privacy, multiple organizational servers adopt FL manner to collaboratively train the prediction model. However, the two proposed methods are far from reasonable, because only the spatial correlation among intra-organization nodes is evaluated and that of inter-organization ones is ignored, the lack of which leads to low forecasting accuracy.

Different from the above researches, Meng \emph{et al.} proposed a model named cross-node federated graph neural network (CNFGNN) \cite{meng2021cross}.
In this work, each traffic sensor serves as a client, and GNN is utilized to capture the spatial relation at the central server. 
However, in the backpropagation process of GNN, the central server has to communicate with clients for updating parameters, which incurs considerable communication overhead. 
Besides, since the transportation graph composed by clients at each round remains unchanged, the captured spatial correlation among multiple clients barely changes, thus failing to cater for traffic fluctuation.

Given the dynamic changes of traffic flow, our work mainly focuses on designing an online method for dynamically capturing spatio-temporal correlation among traffic nodes in FL framework, which is quite different from the above researches.

\section{Problem Formulation and Preliminary Knowledge}
In this section, we firstly formulate the traffic flow forecasting problem with OL mode in FL framework, and then present the preliminary knowledge.
\subsection{Problem Formulation}
Suppose there is a central server and $N$ traffic nodes. All traffic nodes compose a transportation network, denoted as a directed graph $\mathcal{G}= (\mathcal{S}, \mathcal{E})$. $\mathcal{S}= \{s_{1}, s_{2}, \cdots, s_{N}\}$ represents the node set and $s_n$ denotes the $n$-th traffic node. $\mathcal{E} = \left( {{e_{m,n}}} \right) \in {\mathbb{R}^{N \times N}}$ denotes the adjacency matrix of these nodes.
If $s_m$ is adjacent to $s_n$, $e_{m,n}=1$ (otherwise $e_{m,n}=0$).
Let $\mathcal{A}_n$ represent the set of traffic nodes adjacent to $s_n$, and we define that $s_n \in \mathcal{A}_n$. 
Let $x_{t,n}$ denote the traffic speed observed by $s_n$ at the $t$-th time stamp. Primary descriptions are listed in Table \ref{notation}.

\begin{table}
	\caption{Primary Notations and Definitions in Section III}
	\label{notation}
	\centering
	\begin{tabular}{lll}
		\toprule
		Notation& Definition\\
		\midrule
		$\mathcal{G}$& The transportation network composed by traffic nodes.\\
		
		$s_n$& The $n$-th traffic node.  \\
		$e_{m,n}$& The adjacency between $s_n$ and $s_m$.\\
		$\mathcal{A}_n$& Adjacent node set of $s_n$.\\
		$x_{t,n}$& Traffic speed observed by $s_n$ at the $t$-th time stamp.\\
		$f(\cdot)$& Prediction model.\\
		
		$T$& Historical step.\\
		$F$& Forecasting step.\\
		$N$& Client number.\\
		$X_{t,n}^{T}$& Training speed sequence of $s_n$ at the $t$-th round.\\
		$\hat{X}_{t,n}^{F}$& Predicted values generated by $s_n$ at the $t$-th round.\\
		$l$& Loss function.\\
		$\mathcal{R}$& Maximum global round.\\
		$w_t$& Global model parameters of the $t$-th round.\\
		$REG$& Total regret over all clients.\\
		\bottomrule
	\end{tabular}
\end{table}

In this scenario, traffic nodes are regarded as clients to perform traffic flow forecasting using global model parameters from the central server and then incrementally update local models based on newly-observed speed data. 
The central server aggregates local models uploaded by clients to yield the fresh global model. 

At the $t$-th round, $s_n$ receives the global model parameters $w_t$ from the central server and forecasts traffic speeds of future $F$ time stamps based on traffic speeds of $T$ historical time stamps, which can be formulated as 
\begin{equation}
	\hat X_{t,n}^F = {f}(X_{t,n}^T;{w_t}),
\end{equation}
where $X_{t,n}^T= \left( x_{t-T+1,n}, x_{t-T+2,n}, \dots, x_{t,n}\right)$ denotes the sequence of traffic speeds from $t-T+1$ to $t$ observed by $s_n$, and $\hat{X}_{t,n}^F= \left(\hat{x}_{t+1,n}, \hat{x}_{t+2,n}, \dots, \hat{x}_{t+F,n}\right)$ represents the sequence of predicted traffic speeds from ${t+1}$ to ${t+F}$. 
$f(\cdot)$ denotes the chosen prediction model.

Let $X_{t,n}^F = \left(x_{t+1,n}, x_{t+2,n}, \dots, x_{t+F,n} \right)$ denote the true traffic speeds of the future time stamps $s_n$ observes. Then $s_n$ computes the prediction error as follows:
\begin{equation}
	{l}(X_{t,n}^F,f(X_{t,n}^T;w_t)) = {l}(X_{t,n}^F,\hat X_{t,n}^F;{w_t}),
\end{equation}
where $l$ denotes the loss function, and is adopted to evaluate the bias between the predicted and true values.

Let $\mathcal{R}$ represent the maximum number of global round.
We denote $w^*$ as the optimal model parameters which can be obtained in hindsight after the server seeing the data sequences of all $N$ clients over $\mathcal{R}$ rounds \cite{shalev2011online}. The total prediction regret of $N$ clients can be denoted by $REG$, which evaluates the difference in prediction loss using the actual model parameters and $w^*$ \cite{hoi2021online}. 
$REG$ is formulated as
\begin{equation}
REG = \frac{1}{N} \sum\limits_{t = 1}^\mathcal{R} \sum\limits_{n = 1}^N ({{l}(X_{t,n}^F,\hat X_{t,n}^F;{w_t})} - {{l}(X_{t,n}^F,\hat X_{t,n}^F;{w^*})}).
\end{equation}

The objective of traffic flow forecasting adopting OL manner in FL is to minimize the global regrets over all $N$ clients, i.e., $\min REG.$ Since $w^*$ remains constant, we have to iteratively optimize the optimal model parameters $w_t (1\le t\le \mathcal{T})$, so as to generate the minimum regret.

\subsection{Preliminary Knowledge of Federated Learning with Online Learning Manner}

We integrate the OL manner into the widespread used Federated Averaging (FedAvg) method \cite{mcmahan2017communication}, which is elaborated in Algorithm \ref{ag1}. 
Specifically, at the $t$-th round,
each client concurrently optimizes its locally-owned prediction model via Online Gradient Descent (OGD) (Line 9-10) for $E$ epochs using the newly arriving traffic data.
This update manner differs from traditional offline learning mechanism where the prediction model is pre-trained based on historical traffic data obtained in advance. 
After accomplishing the local training procedure, clients upload their updated local models to the central server.
Then, the central server performs model aggregation and yields the fresh global model based on averaging mechanism (Line 5).
\begin{algorithm}
	\caption{FedAvg with OL manner}
	\label{ag1}
	\LinesNumbered
	\KwIn{Initialized model $w_1$, learning rate $\eta$.}
	\KwOut{The global model $w_{\mathcal{R}+1}$.}
	
	\SetKwFunction{Fexecute}{ClientExecute}
	\textsc{\textbf{ServerExecute:}}\\
	\For {$t = 1, 2, \cdots, \mathcal{R}$}{
		\For {$s_n$ in $\mathcal{S}$}{
			{$w_{t+1,n} \gets $ \Fexecute($w_{t}$, $n$, $t$)}\\
			
		}
		{${w_{t+1}} \gets \frac{1}{{{N_t}}}\sum\nolimits_{{s_n} \in {\mathcal{S}}} {{w_{t+1,n}}} $}\\
	}
	\KwRet $w_{\mathcal{R}+1}$\\
	
	\SetKwProg{Fn}{Function}{:}{}
	\Fn{\Fexecute{$w$, $n$, $t$}}{
	    Make prediction via (1).\\
		\For{$e = 1, 2, \cdots, E$}{
			$w \gets w - \eta \nabla l(X_{t,n}^F,\hat X_{t,n}^F;w)$\\
		}
		\KwRet $w$}
		
\end{algorithm}

\section{Methodology}
\begin{figure}[!htbp]
	\centering
	\includegraphics[width=0.49\textwidth]{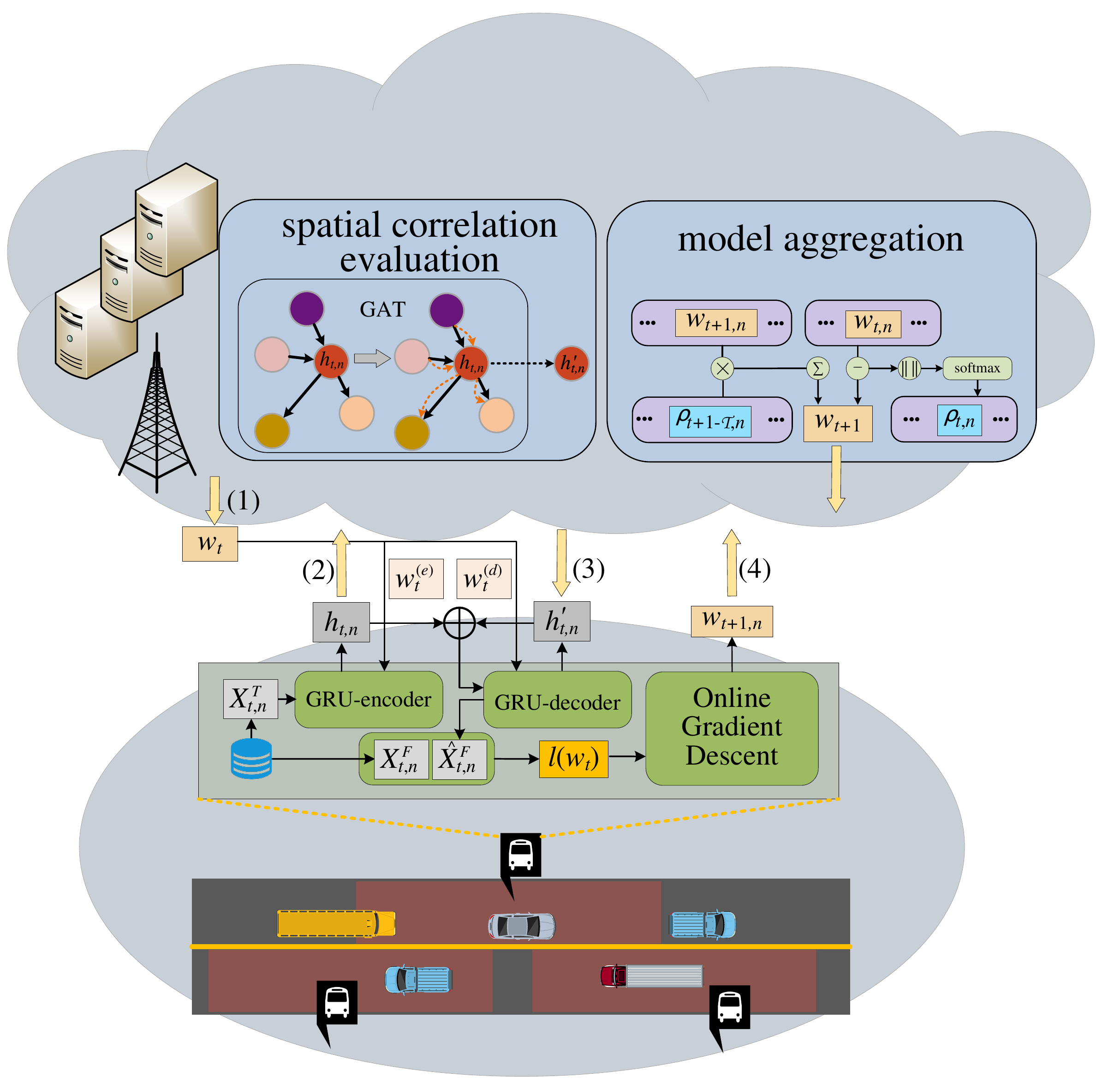}
	\caption{The procedure of FedOSTC. (1) The server distributes the global model. (2) The client uploads the temporal pattern. (3) The server returns the updated temporal pattern. (4) The client uploads local model parameters.
	}
	\label{framework}
\end{figure}

In this section, we provide a detailed introduction of FedOSTC, and the execution process of FedOSTC is elaborated in Fig. \ref{framework}. 
Specifically, we propose a novel spatio-temporal correlation evaluation mechanism which includes obtaining hidden temporal patterns inside traffic speed sequences using GRU-based encoders at clients and assessing spatial correlation among clients at the central server. 
After that, we update local prediction models using OGD algorithm at clients.
Based on the periodic patterns inside traffic flows, we further propose a period-aware aggregation mechanism to generate the fresh global model.
Then, we elaborate the process of the proposed FedOSTC in Algorithm \ref{ag2}.
Last but not least, we analyze the convergence of FedOSTC which theoretically validates the high-performance of FedOSTC.

\subsection{Spatio-Temporal Correlation Evaluation}
An efficient way of assessing spatio-temporal correlation is fundamental to satisfactory prediction performance. Therefore, in this subsection, we elaborates how to evaluate spatial and temporal correlation among clients.
Firstly, clients capture the hidden temporal patterns inside speed sequences respectively with GRU-based encoders. 
Then the central server measures the spatial correlation among clients using GAT and returns the updated hidden temporal patterns to clients.
Finally, clients input the updated hidden patterns into the decoders to generate the predicted results.

\subsubsection{\textbf{Obtaining Temporal Patterns from Speed Sequences at Clients}}
Considering the effectiveness of GRU in processing temporal sequences, we adopt GRU to capture the temporal patterns inside traffic speed sequences.

At the $t$-th round, given $x_{d,n} \in X_{t,n}^F (t-T+1 \le d \le t)$, $s_n$ executes the process of encoding as follows:
\begin{equation}
z_{t,n}^d = sigmoid (W_{t,n}^zx_{d,n}^{} + U_{t,n}^z\hat{h}_{t,n}^{d - 1}),
\end{equation}
\begin{equation}
r_{t,n}^d = sigmoid (W_{t,n}^rx_{d,n}^{} + U_{t,n}^r\hat{h}_{t,n}^{d - 1}),
\end{equation}
\begin{equation}
\tilde h_{t,n}^d = \tanh (W_{t,n}^hx_{d,n} + U_{t,n}^h(r_{t,n}^d \odot \hat{h}_{t,n}^{d - 1})),
\end{equation}
\begin{equation}
\hat{h}_{t,n}^d = (1 - z_{t,n}^d) \odot \hat h_{t,n}^{d - 1} + z_{t,n}^d \odot \tilde h_{t,n}^d,
\end{equation}
where $z_{t,n}^d$, $r_{t,n}^d$, and $\hat{h}_{t,n}^{d}$ represent the update gate, reset gate, and candidate hidden state respectively. $W_{t,n}^z, U_{t,n}^z, W_{t,n}^r, U_{t,n}^r, W_{t,n}^h$, and $U_{t,n}^h$ denote weight matrices and are contained in $w_t^{(e)}$ which denotes the encoder parameters included in $w_t$.

Let $h_{t,n} = \hat{h}_{t,n}^{t}$ denote the hidden temporal pattern learned by the encoder from $X_{t,n}^{T}$.
Then, $h_{t,n}$ is uploaded to the central server for evaluating spatial correlation among clients.

\subsubsection{\textbf{Evaluating Spatial Correlation Among Clients at the Server}}
In the scenario of traffic flow forecasting, due to the closeness of clients' locations, each client is adjacent to some others in the transportation network $\mathcal{G}$. 
Based on this, traffic speed observations at different clients are definitely correlated with each other.
When a certain client makes prediction, historical traffic speed observations of the adjacent clients should be also taken into consideration.
However, since raw traffic data stored at clients cannot be shared for privacy preservation, it is infeasible to capturing spatial correlation among clients using raw data like many centralized methods.

\begin{figure}[!htbp]
	\centering
	\includegraphics[width=0.45\textwidth]{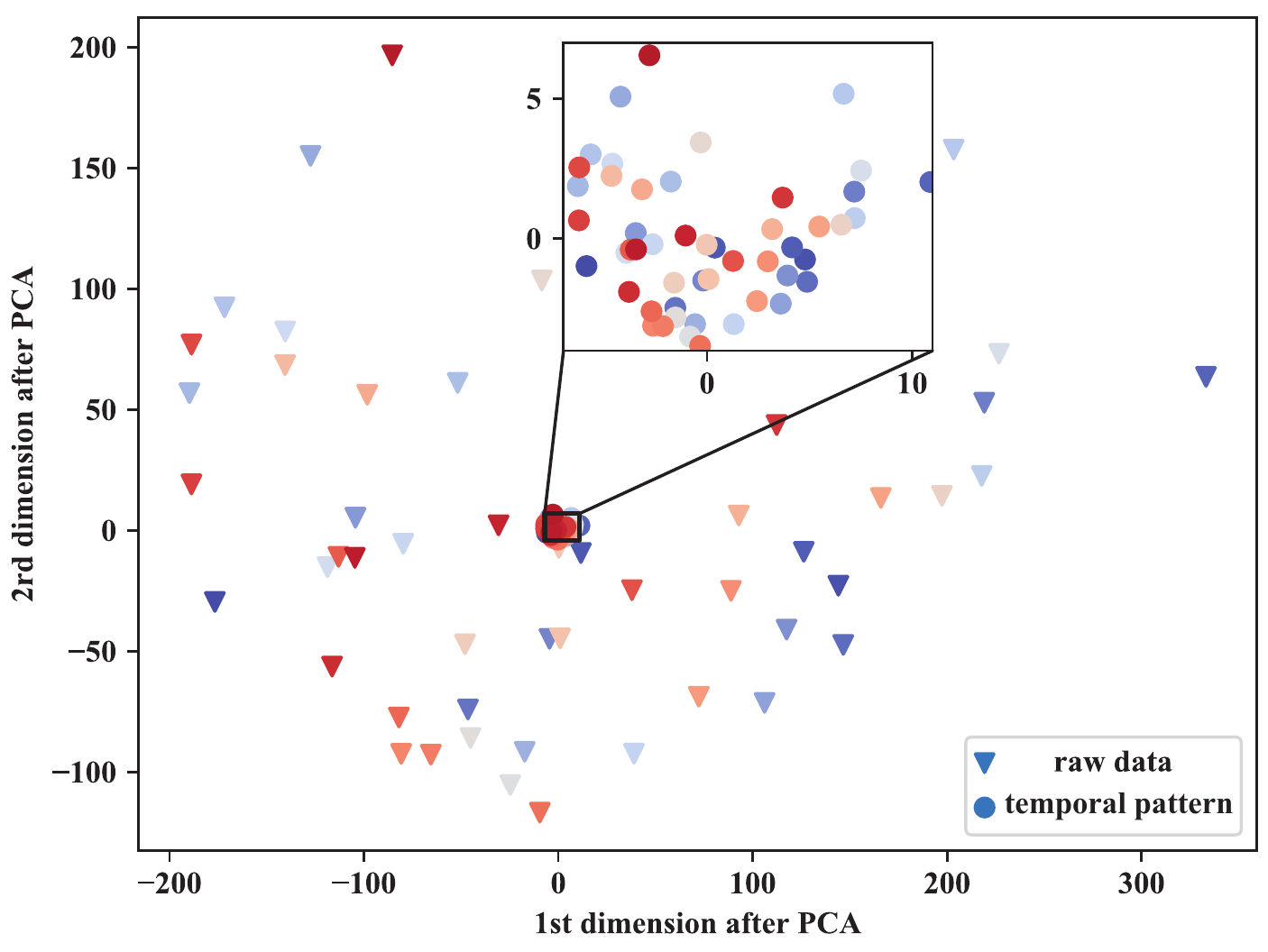}
	\caption{The distributions of raw speed sequences and temporal patterns after dimensional reduction by PCA.}
	\label{data_and_state}
\end{figure}

Inspired by \cite{meng2021cross}, we draw that raw traffic sequences and temporal patterns have similar distribution.
The distributions of temporal patterns and traffic sequences after dimensional reduction by Principal Component Analysis (PCA) are depicted in Fig. \ref{data_and_state}. 
We mark different clients in different colors and can observe that ``circles" and ``triangles" have the same distribution.
Based on this, we utilize temporal patterns output by encoders as the alternative to raw traffic speed for the assessment of spatial correlation among clients.

In this subsection, we evaluate spatial correlation among clients using GAT, where the closeness of spatial relation is assessed by directly calculating attention coefficients of temporal patterns using attention mechanism. 
It has the advantages that attention coefficients eventually express the relation among raw data, thus adapting to the dynamic changes of spatial correlation aroused by traffic fluctuation. 

Specifically, after receiving temporal patterns from clients, the central server firstly calculates the attention scores between two adjacent clients.
The attention score between $s_n$ and $s_m$ at the $t$-th round is denoted as $\xi_{m,n}^t (\xi_{m,n}^t \in \mathbb{R})$, which can be calculated by
\begin{equation}
{\xi_{m,n}^t} = a([{h_{t,n}}||{h_{t,m}}]),\forall {s_m} \in {\mathcal{A}_n},
\end{equation} 
where $[\cdot||\cdot]$ represents the operation of concatenating. 
$a(\cdot)$ is a projection function. 

Furthermore, we denote the attention coefficient between $s_m$ and $s_n$ at the $t$-th round as $\alpha _{m,n}^t$ and it is calculated by
\begin{equation}
{\alpha _{m,n}^t} = \frac{{\exp ({\rm{LeakyReLU}}({\xi_{m,n}^t}))}}{{\sum\nolimits_{s_k \in {\mathcal{A}_n}} {\exp ({\rm{LeakyReLU}}({\xi_{k,n}^t}))} }}.
\end{equation}
It is remarkable that $\alpha _{m,n}^t$ indicates the correlation strength of $s_m$ and $s_n$. 
The larger $\alpha _{m,n}^t$ is, the more related $s_m$ is to $s_n$.

Finally, the central server updates $h_{t,n}$ according to
\begin{equation}
{h'_{t,n}} = sigmoid (\sum\nolimits_{{s_m} \in {\mathcal{A}_n}} {{\alpha_{m,n}^t}{h_{t,m}}} ),
\end{equation}
where ${h'_{t,n}}$ denotes the updated version of $h_{t,n}$. 
It is worth noting that ${h'_{t,n}}$ contains not only the temporal pattern inside $X_{t,n}^{T}$, but the temporal patterns inside speed sequences from adjacent clients. 
Then, the central server transmits ${h'_{t,n}}$ back to $s_n$ for generating predicted values.

The decoder of $s_n$ inputs $h'_{t,n}$ and outputs the predicted results. $w_t^{(d)}$ denotes the decoder parameters and is also included in $w_t$. Each decoder is also a GRU whose execution follows the process of (4)-(7), besides appending a fully-connected layer for making decisions of the final prediction and outputting $\hat{X}_{t,n}^{F}$.

\subsection{Prediction Model Update}
In this subsection, we illustrate the process of prediction model update, which includes incrementally optimizing local models at clients and the period-aware model aggregation mechanism.

\subsubsection{\textbf{Incremental Local Optimization at Clients}}
At the $t$-th round, $s_n$ outputs the predicted result $\hat{X}_{t,n}^{F}$, followed by calculating the prediction loss based on (2).
Then, $s_n$ executes the backpropagation process to compute the current gradient of encoder and decoder.
At the $e$-th local epoch, the model parameters of $s_n$ are optimized using OGD as follows \cite{hoi2021online}:
\begin{equation}
g_{t ,n}^{(e)} = \nabla {l}(X_{t,n}^F,\hat X_{t,n}^F;
w_{t,n}^{(e - 1)}),
\end{equation}
\begin{equation}
w_{t,n}^{(e)} = w_{t,n}^{(e - 1)} - \eta g_{t,n}^{(e)},
\end{equation}
where $g_{t ,n}^{(e)}$ and $w_{t,n}^{(e)}$ denote the gradient and the local model parameters of $s_n$ at the $e$-th epoch in the $t$-th global round respectively.  
$\eta$ represents the learning rate and $w_{t,n}^{(0)} = {w_t}$. 
$s_n$ executes the above process for $E$ epochs, and then transmits the up-to-date local model parameters $w_{t+1,n}$ ($w_{t+1,n}=w_{t,n}^{(E)}$) to the central server for aggregation. 

\subsubsection{\textbf{Period-Aware Aggregation Mechanism at the Server}}
In traditional FL methods, such as FedAvg \cite{mcmahan2017communication}, the server aggregates local model parameters, and generates the fresh global model $w_{t+1}$ at the $t$-th round as
\begin{equation}
{w_{t+1}} = \frac{1}{N}\sum\limits_{n = 1}^N {{w_{t+1,n}}.} 
\end{equation}

It is plain that $w_{t+1}$ is updated based on traffic data at the $t$-th round.  
Since traffic sequences at different rounds have variant distributions, this aggregation mechanism makes $w_{t+1}$ more applicable to the traffic data of the $t$-th round, compared with $w_t$. 
Therefore, if the server treats $w_{t+1}$ as the initial model parameters at beginning of the $t$-th round, the prediction accuracy can be increased.
However, it is far from reasonable, since $w_{t+1}$ is generated in the end of the $t$-th round.

\begin{figure}[!htbp]
	\centering
	\includegraphics[width=0.5\textwidth]{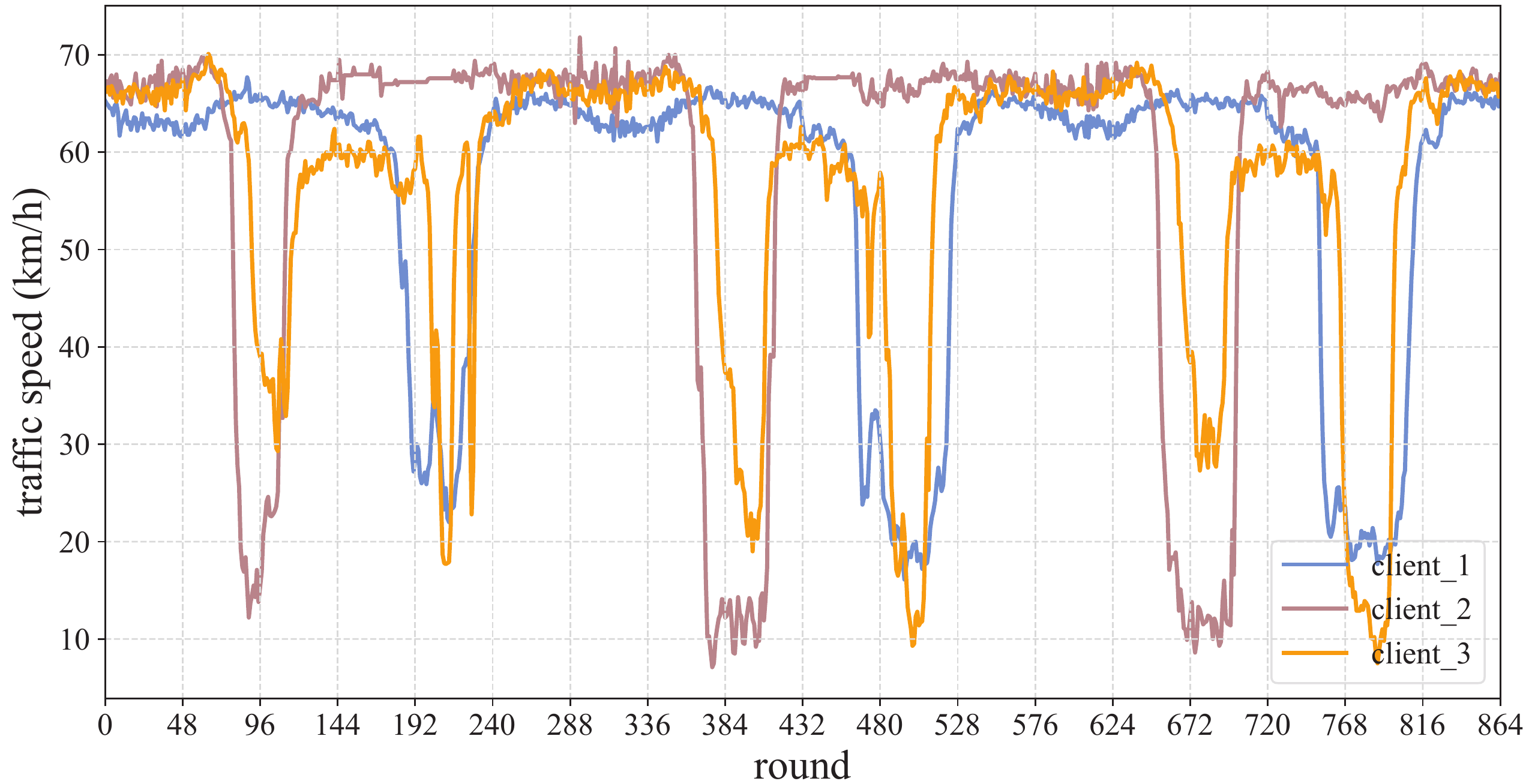}
	\caption{Raw traffic flows of three clients in 5 days.}
	\label{traffic_speed}
\end{figure}

Furthermore, we discover that each traffic flow has the same built-in period. 
As shown in Fig. \ref{traffic_speed}, each client has similar distribution every 288 time stamps (clients observe traffic speeds every 5 minutes and therefore one day is split into 288 time stamps), except for a few ones.
Let $\mathcal{T}$ denote the period of traffic flows. 
It is reasonable to deem that local model parameters $w_{t,n}$ and global model parameters $w_{t+1}$ are similar with $w_{t+\mathcal{T},n}$ and $w_{t+\mathcal{T}+1}$ respectively.

Inspired by this, at the $t$-th round, we calculate the correlation coefficients between previous local models $w_{t,n} (\forall s_n \in \mathcal{S})$ and the fresh global model $w_{t+1}$ as
\begin{equation}
{\rho _{t,n}} = \frac{\exp (-||w_{t,n} - w_{t+1}||)}{\sum\nolimits_{m = 1}^N {\exp (-||w_{t,m} - w_{t+1}||)}}, \forall s_n \in \mathcal{S},
\end{equation}
where $\rho_{t,n}$ denotes the correlation coefficient between $w_{t,n}$ and $w_{t+1}$. 
If the server performs weighted aggregation based on $\rho_{t,n}$ in the end of the ($t-1$)-th round as $\sum\nolimits_{{s_n} \in {\mathcal{S}}} \rho_{t,n} w_{t,n}$, the newly-generated global model $w_t$ will apply to the data sequences of the $t$-th round better than the original $w_t$ yielded using averaging mechanism as (13).
Furthermore, $\rho_{t,n}$ also enables to approximate the similarity between $w_{t+\mathcal{T},n}$ and $w_{t+\mathcal{T}+1}$, and thus it is directly adopted as the aggregation weight of $s_n$ at the $(t+\mathcal{T}-1)$-th round.

Specifically, the execution process of the proposed period-aware aggregation mechanism is as follows.
\begin{itemize}
    \item {If $t \le \mathcal{T}$, the central server firstly generates the fresh global model $w_t$ as (13) and then calculates correlation coefficients as (14).
    }
    \item {If $t > \mathcal{T}$, the central server calculates correlation coefficients as (14), following yielding $w_{t+1}$ as
    \begin{equation}
    {w_{t+1}} = \sum\limits_{n = 1}^N {{\rho _{t+1-\mathcal{T},n}}}  \cdot {w_{t+1 ,n}}.
    \end{equation}
    }
\end{itemize}

It is reasonable to deem that the averaging aggregation mechanism formulated in (13) is a special case of the weighted aggregation in (15), with $\rho_{t+1-\mathcal{T},n}=\frac{1}{N},\forall{s_n \in \mathcal{S}}$.

By this means, the generated fresh global model can well adapt to the subsequent traffic data better, which yields more satisfactory prediction performance.

\begin{algorithm}
	\caption{FedOSTC}
	\label{ag2}
	\LinesNumbered
	\KwIn{initialized global model $w_1$, traffic flow sequence [$x_{t,n}^{T}$, $x_{t,n}^{F}$], maximum global round $R$, maximum local epoch $E$.}
	\KwOut{The global model $w_{R+1}$}
	
	\SetKwFunction{Fexecute}{ClientExecute}
	\SetKwFunction{Fencode}{ClientEncode}
	\SetKwFunction{Fdecode}{ClientDecode}
	
	\textsc{\textbf{ServerExecute:}}\\
	\For {$t = 1, 2, \cdots, \mathcal{R}$}{
	    \For {$s_n \in \mathcal{S}$}{
	        $w_{t+1,n} \gets$ \Fexecute($w_t$, $n$, $t$)\\
	    
	    \If {$t \le \mathcal{T}$}{
	        Generate $w_{t+1}$ via (13).\\
	    }
	    \Else{
	        Generate $w_{t+1}$ via (15).\\
	    }
	    Calculate $\rho_{t,n}$ via (14).\\}
	}
	\KwRet $w_{\mathcal{R}+1}$.\\
	\SetKwProg{Fn}{Function}{:}{}
	\Fn{\Fexecute{$w_t$, $n$, $t$}}{
	    [$w_{t,n,e}^{(0)}$, $w_{t,n,d}^{(0)}$] = $w_{t,n}^{(0)}$ = $w_t$.\\
	    \For {$e = 1, 2, \cdots, E$}{
	        $h_{t,n}$ $\gets$ \Fencode($x^{T}_{t,n}$, $w_{t,n,e}^{(e)}$, $n$).\\
	        \If{$e = 1$}{
	            Get $h^{'}_{t,n}$ via (8)-(10).\\
	        }
	        $\hat{x}^{F}_{t,n} \gets $ \Fdecode($h_{t,n}$, $h^{'}_{t,n}$, $w_{t,n,d}^{(e)}$, $n$).\\
	        $w_{t,n}^{(e)}$ = [$w_{t,n,e}^{(e)}$, $w_{t,n,d}^{(e)}$].\\
	        $loss$ $\gets$ ${l}(X_{t,n}^F,\hat X_{t,n}^F;w_{t,n}^{(e)}$)\\
	        Update $w_{t,n}^{(e)}$ via (11)(12).\\
	    }
	    $w_{t+1,n}$ = $w_{t,n}^{(E)}$ = [$w_{t,n,e}^{(E)}$, $w_{t,n,d}^{(E)}$].\\
	    \KwRet $w_{t+1,n}$.\\
	}
	
	\SetKwProg{Fn}{Function}{:}{}
	\Fn{\Fencode{$x^{T}_{t,n}$, $w_{t,n,e}^{(e)}$, $n$}}{
	    With $x_{t,n}^T$ input, get $h_{t,n}$ through (4)-(7).\\
	    \KwRet $h_{t,n}$\\
	}
	\SetKwProg{Fn}{Function}{:}{}
	\Fn{\Fdecode{$h_{t,n}$, $h^{'}_{t,n}$, $w_{t,n,d}^{(e)}$, $n$}}{
	    Get $\hat{x}^{F}_{t,n}$ from  GRU.\\
	    \KwRet $\hat{x}^{F}_{t,n}$.
	}
\end{algorithm}

\subsection{\textbf{Execution Process of FedOSTC}}
Algorithm \ref{ag2} elaborates the process of FedOSTC, where $w_{t,n,e}^{(e)}$ and $w_{t,n,d}^{(e)}$ denote the parameters of encoder and decoder on $s_n$ at the $e$-th epoch in the $t$-th round respectively.
At each global round, the server transmits current global model to clients. 
Encoders at clients input traffic sequences and output the temporal patterns of these traffic sequences (Line 23-25). 
The server evaluates the spatial correlation among clients using GAT, and then updates hidden states from clients (Line 16). 
When clients receive the updated hidden states from the server, they make prediction, calculate prediction loss and optimize parameters of encoders and decoders via OGD, which is iterated for $E$ epochs (Line 17-20). 
Then clients transmit their updated model parameters to the central server for aggregation using the period-aware weighted aggregation mechanism (Line 5-9). 
The process continues until the maximum round $\mathcal{R}$ arrives.

\subsection{Convergence Analysis of FedOSTC}
In this subsection, the convergence analysis of FedOSTC is demonstrated theoretically. Based on the theory of Online Convex Optimization (OCO) \cite{shalev2011online}, we obtain the upper bound of $REG$.

For simplicity, we reformulate ${l}(X_{t,n}^F,\hat X_{t,n}^F;{w_t})$ as $l(w_t)$, since the loss and prediction results are dependent on the adopted model parameters.
Firstly, some assumptions are introduced.
\newtheorem{assumption}{Assumption}
\begin{assumption}
	The loss function $l(\cdot)$ is L-smooth, i.e., $l(x) - l(y) \le \left\langle {\nabla l(y),x - y} \right\rangle  + \frac{L}{2}||x - y|{|^2}, \forall x,y, \exists L > 0.$
\end{assumption}

\begin{assumption}
	The gradient of loss function $l(\cdot)$ is uniformly bounded, i.e., ${\rm \mathbb{E}}(||\nabla {l}( \cdot )|{|^2}) \le {G^2}.$
\end{assumption}

\begin{assumption}
	REG is bounded below, i.e., $RE{G_{\min }} = REG{^*} >  - \infty.$
\end{assumption}

To Facilitate the analysis on the convergence rate of FedOSTC, the period-aware weighted aggregation mechanism described in (15) is transformed.
\newtheorem{lemma}{Lemma}
\begin{lemma}
	The aggregation mechanism described in (15) can be reformulated as
	\begin{equation}
	{w_{t + 1}} = {w_t} - \eta \sum\limits_{n = 1}^N {\sum\limits_{e = 1}^E {{\rho _{t,n}} \cdot g_{t,n}^{(e)}} } .
	\end{equation}
\end{lemma}
Therefore, the dynamic weighting aggregation mechanism can be considered as a relational expression of $w_t$ and $w_{t+1}$. The proof of {\scshape Lemma} 4.1 is elaborated as follows.
\begin{proof}
    With (11) and (12), we can get
    \begin{align}
        w_{t,n}^{(E)} &= w_{t,n}^{(E - 1)} - \eta g_{t,n}^{(E)}
        = w_{t,n}^{(E - 2)} - \eta  g_{t,n}^{(E - 1)} - \eta  g_{t,n}^{(E)}
        \notag
        \\
        &= \cdots = w_{t,n}^{(0)} - \eta g_{t,n}^{(1)} -  \cdots  - \eta g_{t,n}^{(E - 1)} - \eta g_{t,n}^{(E)}
        \notag
        \\
        &= {w_t} - \eta \sum\limits_{e = 1}^E {g_{t,n}^{(e)}}.
    \end{align}
	Hence, we can complete the proof as
	\begin{align}
	    {w_{t + 1}} &= \sum\limits_{n = 1}^N {{\rho _{t,n}} {w_{t + 1,n}}} = \sum\limits_{n = 1}^N {{\rho _{t,n}}({w_t} - \eta \sum\limits_{e = 1}^E {g_{t,n}^{(e)}} )}
	    \notag
	    \\
	    &= \sum\limits_{n = 1}^N {{\rho _{t,n}}{w_t} - \eta \sum\limits_{n = 1}^N {\sum\limits_{e = 1}^E {{\rho _{t,n}} g_{t,n}^{(e)}} } }
	    \notag
	    \\
	    &= {w_t} - \eta \sum\limits_{n = 1}^N {\sum\limits_{e = 1}^E {{\rho _{t,n}} g_{t,n}^{(e)}} }.
	\end{align}
\end{proof}
\newtheorem{theorem}{Theorem}
\begin{theorem}
	With {\scshape Assumption} 1-3 and {\scshape Lemma} 1 held, after $\mathcal{R}$ round, $REG$ in FedOSTC has an upper bound: 
	\begin{equation}
	{\rm \mathbb{E}}[REG_{\min}] \le {\rm{(1}} + EN + \frac{L}{2}EN{\rm{)}}\eta \mathcal{R}{G^2}.
	\end{equation}
\end{theorem}
{\scshape Theorem} 4.2 indicates that the minimum regret of FedOSTC has an upper bound. Furthermore, $REG_{\min}=o(\mathcal{R})$ indicates that FedOSTC performs as well as the the optimal parameters $w^*$. Therefore, FedOSTC is efficient in tackling with the issue of traffic flow forecasting. The proof of {\scshape Theorem} 4.2 is detailed.

\begin{proof}
	Since $l$ is L-smooth, we can transform {\scshape Assumption} 1 as
	\begin{align}
	    & {l}({w_t}) - {l}({w_{t + 1}})
	   \le \left\langle {\nabla {l}({w_{t + 1}}),{w_t} - {w_{t + 1}}} \right\rangle  + \frac{L}{2}||{w_t} - {w_{t + 1}}|{|^2}
	   \notag
	   \\
	   &= \left\langle {\nabla {l}({w_{t + 1}}),\eta \sum\limits_{e = 1}^E {\sum\limits_{n = 1}^N {{\rho _{t,n}}g_{t,n}^{(e)}} } } \right\rangle 
	   {\rm{   }} + \frac{L}{2}||\eta \sum\limits_{e = 1}^E {\sum\limits_{n = 1}^N {{\rho _{t,n}}g_{t,n}^{(e)}} } |{|^2}
	   \notag
	   \\
	   &\le \eta ||\nabla {l}({w_{t + 1}}) - \sum\limits_{e = 1}^E {\sum\limits_{n = 1}^N {{\rho _{t,n}}g_{t,n}^{(e)}} } |{|^2}
	   {\rm{   }} + \frac{L}{2}||\eta \sum\limits_{e = 1}^E {\sum\limits_{n = 1}^N {{\rho _{t,n}}g_{t,n}^{(e)}} } |{|^2}
	   \notag
	   \\
	   &\le \eta ||\nabla {l}({w_{t + 1}})|{|^2} + (1+\frac{L}{2})\eta \sum\limits_{e = 1}^E {\sum\limits_{n = 1}^N {||{\rho _{t,n}}g_{t,n}^{(e)}} } |{|^2}
	\end{align}
	With {\scshape Assumption} 2, the inequation above can be reformulated as
	\begin{align}
	    {\mathbb{E}}[{l}({w_t}) - {l}({w_{t + 1}})]
	    &\le \eta {\rm \mathbb{E}}||\nabla {l}({w_{t + 1}})|{|^2} 
            \notag
            \\
            & + (1+\frac{L}{2})\eta \sum\limits_{e = 1}^E {\sum\limits_{n = 1}^N {{\rm \mathbb{E}}||{\rho _{t,n}}g_{t,n}^{(e)}} } |{|^2}
	    \notag
	    \\
	    &\le \eta {G^2} + (1+\frac{L}{2})\eta EN{G^2}.
	\end{align}
	With {\scshape Assumption} 3, the minimum regret can be obtained as
	\begin{align}
	    {\mathbb{E}}[REG_{\min}]
	    &= {\rm \mathbb{E}}\left\{ {\frac{1}{N}\sum\limits_{n = 1}^N {\sum\limits_{t = 1}^{\mathcal{R}} {[{l}({w_t}) - {l}({w^*})]} } } \right\}
            \notag
            \\
	   &\le {\rm \mathbb{E}}\left\{ {\frac{1}{N}\sum\limits_{n = 1}^N {\sum\limits_{t = 1}^{\mathcal{R}} {[{l}({w_t}) - {l}({w_{t + 1}})]} } } \right\}
	   \notag
	   \\
	   &\le \frac{1}{N}\sum\limits_{n = 1}^N {\sum\limits_{t = 1}^{\mathcal{R}} {(\eta {G^2} + \eta EN{G^2} + \frac{L}{2}\eta EN{G^2})} }
	    \notag
	    \\
	    &= (1+EN + \frac{L}{2}EN{\rm{)}}\eta {\mathcal{R}}{G^2}.
	\end{align}
\end{proof}

\section{Experiments}
\begin{table*}[h]
  \centering
  \caption{Comparison in Prediction Performance of Six Methods on Two Datasets with Different Forecasting Steps}
  \renewcommand\arraystretch{1.4}
  \small
    \begin{tabular}{c||cc|cc|cc}
    \hline
    \multirow{3}[6]{*}{Methods} & \multicolumn{6}{c}{PEMS-BAY}\\
\cline{2-7}          & \multicolumn{2}{c|}{5min ($F=1$)} & \multicolumn{2}{c|}{30min ($F=6$)} & \multicolumn{2}{c}{1h ($F=12$)} \\
\cline{2-7} & RMSE  & MAE & RMSE  & MAE & RMSE  & MAE \\
    \hline
    CenterOff & 1.898  & 1.019  & 3.602  & 1.736  & 4.693  & 2.255 \\
    CenterOn & 1.614  & \underline{1.007}  & 2.243  & 1.353  & \underline{2.327}  & \underline{1.371}  \\
    FedAvgOff & 1.802       & 1.023      & 3.717  & 1.784  &  5.253     &   2.406    \\
    FedAvgOn & \textbf{0.996}  & \textbf{0.996}  & \underline{1.952}  & \underline{1.712}  & 2.661  & 2.306\\
    CNFGNN &  1.691     &  0.983     & 3.758  & 1.882  & 5.265  & 2.523   \\
    \hline
    FedOSTC & \underline{1.014}  & 1.014  & \textbf{1.450}  & \textbf{1.209}  & \textbf{1.601}  & \textbf{1.254}  \\
    \hline
    $\uparrow$ & -0.108 & -0.018 & +0.502 & +0.503 & +0.726 & +0.117\\
    \hline
    \hline
    \multirow{3}[6]{*}{Methods} & \multicolumn{6}{c}{METR-LA} \\
\cline{2-7}          & \multicolumn{2}{c|}{5min ($F=1$)} & \multicolumn{2}{c|}{30min ($F=6$)} & \multicolumn{2}{c}{1h ($F=12$)} \\
\cline{2-7} & RMSE  & MAE & RMSE  & MAE & RMSE  & MAE \\
    \hline
    CenterOff & 5.796  & 3.248  & 7.975  & 4.392  & 9.346  & 5.184  \\
    CenterOn & 4.949  & 3.261  & 5.889  & \underline{3.639}  & 6.137  & \underline{3.673}  \\
    FedAvgOff &   5.843    &   3.321    & 8.163  & 4.445  & 9.522  & 5.223  \\
    FedAvgOn   & \textbf{2.945}  & \textbf{2.945}  & \underline{4.588}  & 3.877  & \underline{5.714}  & 4.744  \\
    CNFGNN &  5.705     &  3.157     & 8.160  & 4.536  & 9.523  & 5.242 \\
    \hline
    FedOSTC  & \underline{3.006}  & \underline{3.006}  & \textbf{4.129}  & \textbf{3.353}  & \textbf{4.598}  & \textbf{3.484}  \\
    \hline
    $\uparrow$  & -0.061 & -0.061 & +0.459 & +0.286 & +1.116 & +0.189\\
    \hline
    \end{tabular}%
  \label{resultes}%
\end{table*}%
In this section, comprehensive experiments are conducted to validate the effectiveness of our proposed method FedOSTC. Firstly, the system configuration is introduced. Then, the prediction performance of FedOSTC is compared with that of five baselines. 
Furthermore, we compare the generalization ability of the global models in the FL-based methods. Finally, additional experiments are conducted to explore the effect of local epoch on prediction performance.

\subsection{System Configuration}
\subsubsection{\textbf{Datasets and Metrics}}
The experiments are conducted on two real-world datasets, i.e., PEMS-BAY and METR-LA respectively\cite{li2017diffusion}. PEMS-BAY contains traffic speed from 01/01/2017 to 31/05/2017 collected by 325 sensors. METR-LA contains speed information from 207 sensors ranging from 01/03/2012 to 30/06/2012. In the two datasets, sensors observe traffic speed every 5 minutes. We randomly select 50 sensors as clients and utilize the speed data from Sunday to Thursday in May for experiments. The adjacency matrixes of clients for the two datasets are constructed like \cite{meng2021cross}. 
Root Mean Square Error (RMSE) and Mean Absolute Error (MAE) are adopted to evaluate the prediction error.

\subsubsection{\textbf{Experiment Setting}}
FedOSTC is implemented with PyTorch and all experiments are conducted on a server with an Intel(R) Xeon(R) Gold 6230 CPU and eight Nvidia Tesla V100S-PCIe GPUs. In all experiments, the encoder and decoder are both a GRU layer with 64 and 128 cells respectively. $T$, $E$, and $\eta$ are set to 12, 5, and 0.001 respectively. The settings of other methods are the same as the respective literature, unless mentioned otherwise.

\subsection{Comparisons of Forecasting Performance}
\begin{figure*}[!htbp]
	\centering
	\includegraphics[width=1.0\textwidth]{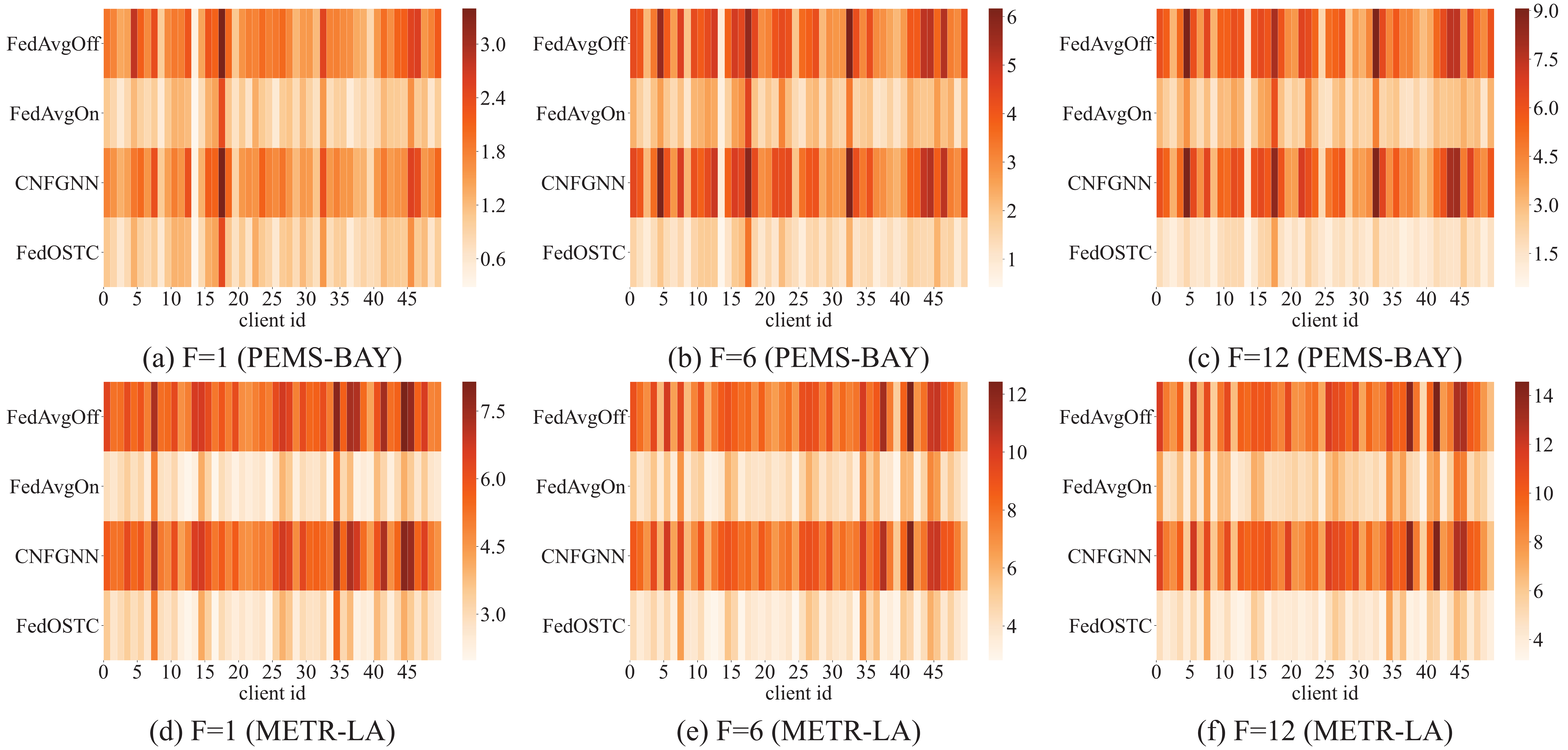}
	\caption{Prediction RMSEs of clients.}
	\label{rmse}
\end{figure*}

\subsubsection{\textbf{Baseline Methods}}
We compare FedOSTC with five baseline methods described as follows.
\begin{itemize}
	\item {
	\textbf{CenterOff}\cite{cho2014learning}: The central server trains the prediction model using traffic data uploads from traffic nodes by batch learning manner. Spatial correlation among traffic nodes is not evaluated.
	}
	\item{
	\textbf{CenterOn}: The execution process is the same as CenterOff except for that the central server online updates the prediction model.
	}
	\item{
	\textbf{FedAvgOff}\cite{mcmahan2017communication}: In the FedAvg method, clients update the prediction model based on batch learning manner and the server aggregates updated local models via averaging mechanism. Spatial correlation among clients is not evaluated.}
	\item{
	\textbf{FedAvgOn}: It is the same as FedAvgOff, except for that clients update the prediction model based on OL manner.}
	\item{
	\textbf{CNFGNN}\cite{meng2021cross}: The execution process is like FedAvgOff, except for that the server evaluates spatial dependence using GNN.}

\end{itemize}

The prediction errors of FedOSTC and the baselines on two datasets with different forecasting steps are presented in Table \ref{resultes}. Thereinto, `$\uparrow$' denotes the performance gains of FedOSTC over the best-performed baseline.

If the forecasting step is shorter ($F$ is lower), the speed sequence to be forecasted ($X_{t,n}^F$) is more likely to have similar distribution with the historical speed sequence ($X_{t,n}^T$). This is the reason why prediction errors increase with the longer forecasting step. 
When $F=1$, the performance gains of FedOSTC are negative and the absolute values are so low as to be ignored. 
However, when $F=6$ and 12, FedOSTC performs best and generates considerable performance gains.

In terms of both RMSE and MAE, CenterOn and FedAvgOn outperform CenterOff and FedAvgOff respectively, which indicates the high-efficiency of OL manner in the task of traffic flow forecasting. It is because that the prediction model is incrementally optimized based on fine-grained speed data, which can sufficiently obtain the dynamic changes in speed data.

In the three methods adopting batch learning manner, CenterOff achieves the best prediction performance on both datasets. It shows that the centralized method (CenterOff) perform better in evaluating spatio-temporal correlation among clients than the FL methods (FedAvgOff and CNFGNN). However, our proposed FedOSTC performs best among the three methods adopting OL manner, which indicates the effectiveness of spatio-temporal correlation mechanism and period-aware weighted aggregation mechanism. 

\begin{figure*}[!htbp]
	\centering
	\includegraphics[width=1\textwidth]{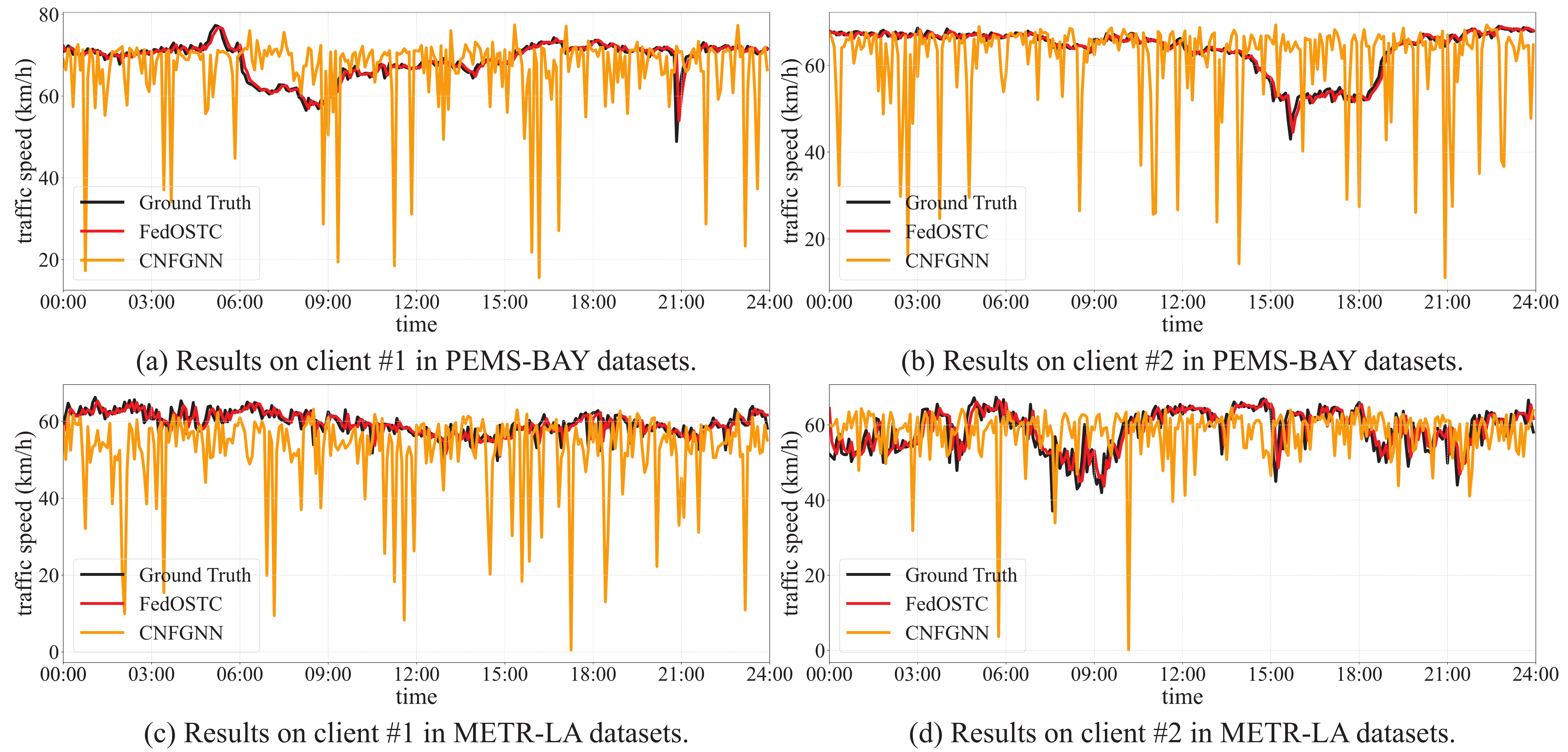}
	\caption{Ground truth and forecasting values of CNFGNN and FedOSTC.}
	\label{y_and_y_pred}
\end{figure*}

To further evaluate the high-efficiency of FedOSTC in traffic flow forecasting, ground truth values and prediction results of FedOSTC and CNFGNN are illustrated in Fig. \ref{y_and_y_pred}.
It is explicit that FedOSTC yields much smaller deviation, compared with CNFGNN, which is the existing best FL method. Furthermore, the distribution of prediction values from FedOSTC is similar to that of ground truth values.

\begin{figure*}[!htbp]
	\centering
	\includegraphics[width=1.0\textwidth]{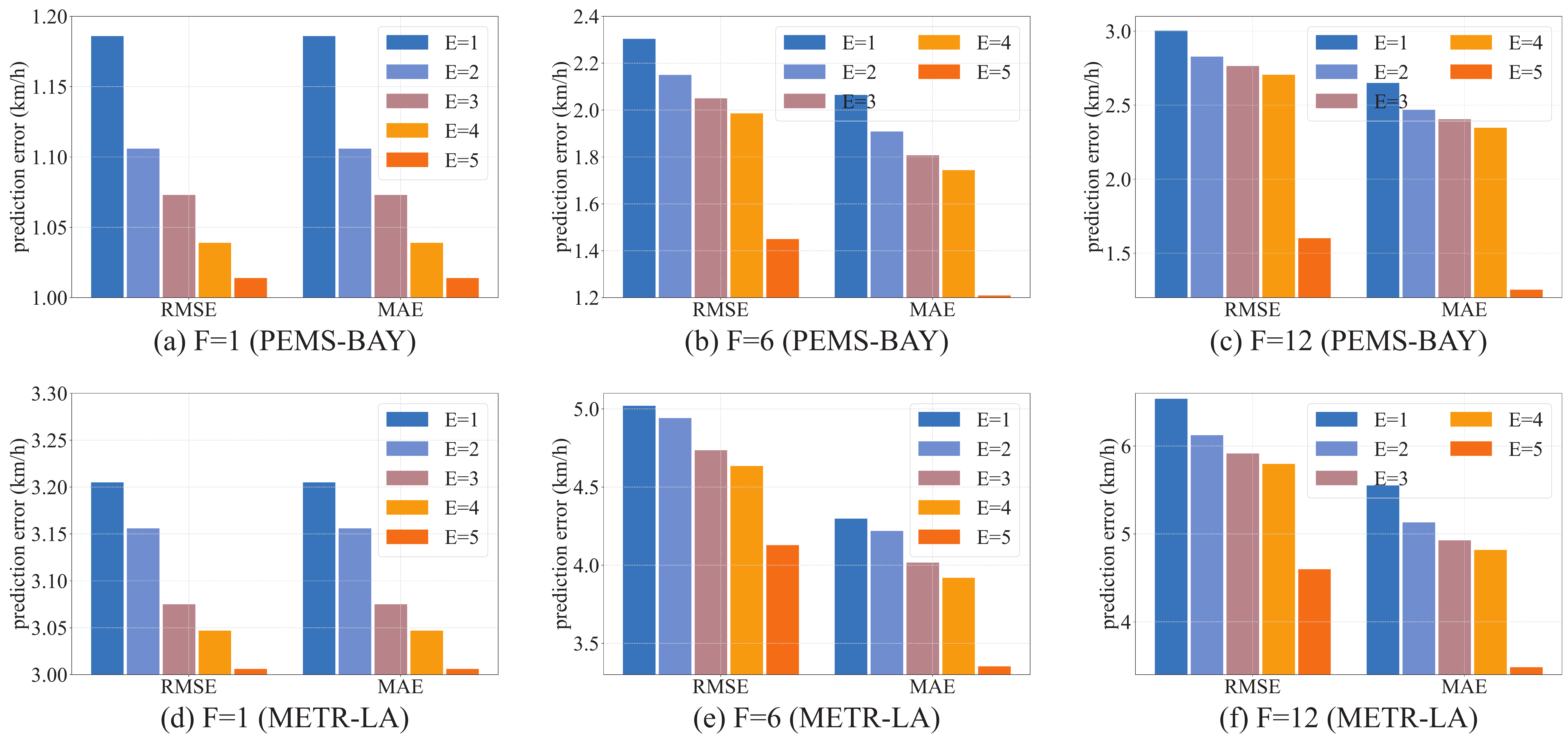}
	\caption{Local epoch versus prediction errors.}
	\label{epoch}
\end{figure*}
\subsection{Comparisons of Generalization Ability}
In the scenario of traffic flow forecasting, the observed traffic speeds from clients have different distribution. While in FL framework, all clients train a same prediction model cooperatively. We hope the global prediction model from the central server has good generalization ability so that all clients can achieve satisfactory performance. Therefore, in this subsection, we compare model generalization ability of the four methods based on FL framework. The prediction RMSEs of clients on two datasets with different forecasting steps are illustrated in Fig.\ref{rmse}. 
By comparing FedAvgOff and CNFGNN with FedAvgOn and FedOSTC respectively, we can observe that RMSE variance of OL manner is much smaller than that of batch learning manner, which indicates that OL manner can yield global model with better generalization ability in FL. 
Furthermore, the prediction RMSE of FedOSTC is lower than that of FedAvgOn for almost all clients. The RMSE variance over all clients in FedOSTC is obviously lower than that in FedAvgOn. Therefore, by adopting the proposed spatio-temporal correlation evaluating mechanism and period-aware weighted aggregation mechanism, FedOSTC can increase the generalization ability of global model.

\subsection{Effect of E on Prediction Performance}
In the scenario of traffic flow forecasting, clients make prediction and perform incremental optimization based on only a speed sequence. It is quite different from those methods adopting batch learning manner, where the local batch size is set to 128. 
In this subsection, we explore the effect of local epoch on prediction performance and further consider whether 5 epochs are redundant for incremental optimization.
The prediction RMSEs and MAEs on two datasets with different settings of $E$ are illustrated in Fig. \ref{epoch}.

We observe that different settings of $E$ yield various effects on prediction errors. On the two datasets, both prediction RMSEs and MAEs get lower with more local epochs, which indicates that the internal patterns of speed sequences can be obtained better with more local optimizations. On the other hand, larger local epoch means clients have to spare more computation resources for local optimization, which poses much burden on the resource-limited clients. Therefore, in the future research, we will explore the balance between local epoch and prediction performance.

\section{Conclusion}
In this paper, we proposed a novel method named FedOSTC for traffic flow forecasting. 
To boost prediction performance, we dynamically assessed spatio-temporal correlation among clients.
Specifically, the temporal patterns inside traffic data sequences are obtained at clients.
Since spatial correlation among clients varies with traffic fluctuation, the spatial dependence of the observed traffic data is assessed dynamically via GAT.
Given the periodic changes of traffic flows, we proposed a period-aware aggregation mechanism to aggregate the online-updated local models, aiming to improve the generalization of the fresh global model for the subsequent traffic data.
Last but not least, we conducted extensive experiments on METR-LA and PEMS-BAY datasets to verify the effectiveness of OL manner in traffic flow forecasting and the superiority of FedOSTC, compared with state-of-the-art methods.

Given the observation that larger local epochs yield higher prediction accuracy but more computation sources, we will explore the balance between computation overhead and prediction performance in the future research and refine FedOSTC from the aspects of reducing communication overhead, decreasing forecasting time, etc.


\bibliographystyle{IEEEtran}
\bibliography{temp}

\newpage
\begin{IEEEbiography}[{\includegraphics[width=1in,height=1.25in,clip,keepaspectratio]{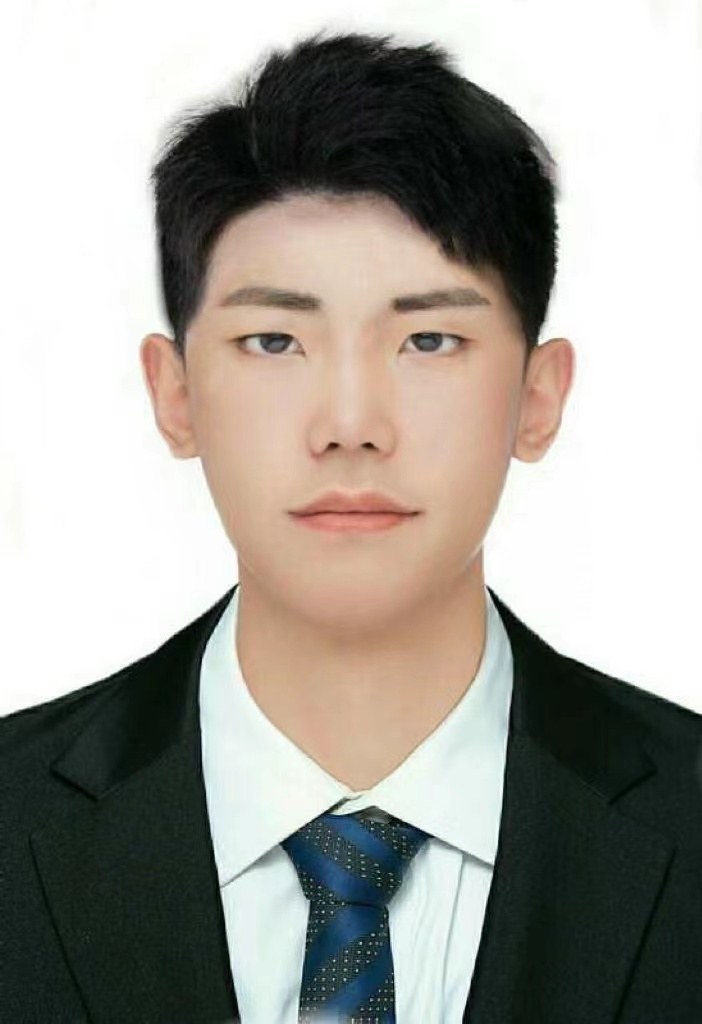}}]{Qingxiang Liu}
	is currently pursuing the Ph.D degree in University of Chinese Academy of Sciences, Beijing, China, with the Institute of Computing Technology, Chinese Academy of Sciences, Beijing, China. 
	He has authored or coauthored several papers at IEEE Transactions on Computational Social Systems, Future Generation Computer Systems and so on. His current research interests include federated learning and edge computing. 
\end{IEEEbiography}
\vspace{-1cm}
\begin{IEEEbiography}[{\includegraphics[width=1in,height=1.25in,clip,keepaspectratio]{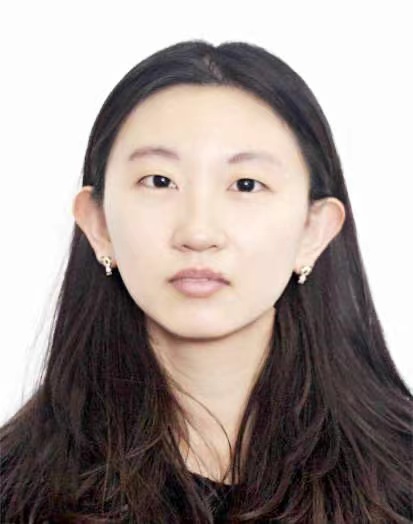}}]{Sheng Sun}
 received her B.S. and Ph.D degrees in computer science from Beihang University, China, and the University of Chinese Academy of Sciences, China, respectively. She is currently an assistant professor at the Institute of Computing Technology, Chinese Academy of Sciences, Beijing, China. Her current research interests include federated learning, mobile computing and edge intelligence. 
\end{IEEEbiography}
\vspace{-1cm}
\begin{IEEEbiography}[{\includegraphics[width=1in,height=1.25in,clip,keepaspectratio]{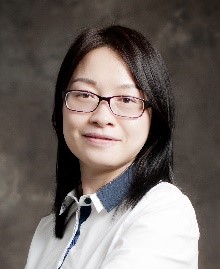}}]{Min Liu}
  (Senior Member, IEEE) received her B.S. and M.S. degrees in computer science from Xi’an Jiaotong University, China, in 1999 and 2002, respectively. She got her Ph.D in computer science from the Graduate University of the Chinese Academy of Sciences in 2008. She is currently a professor at the Networking Technology Research Centre, Institute of Computing Technology, Chinese Academy of Sciences. Her current research interests include mobile computing and edge intelligence.
\end{IEEEbiography}
\vspace{-1cm}
\begin{IEEEbiography}[{\includegraphics[width=1in,height=1.25in,clip,keepaspectratio]{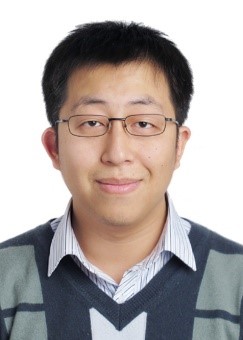}}]{Yuwei Wang}
  received his Ph.D. degree in computer science from the University of Chinese Academy of Sciences, Beijing, China. He is currently a Senior Engineer (equivalent to Associate Professor) at the Institute of Computing Technology, Chinese Academy of Sciences, Beijing, China. His current research interests include federated learning, mobile edge computing, and next-generation network architecture.
\end{IEEEbiography}
\vspace{-1cm}
\begin{IEEEbiography}[{\includegraphics[width=1in,height=1.25in,clip,keepaspectratio]{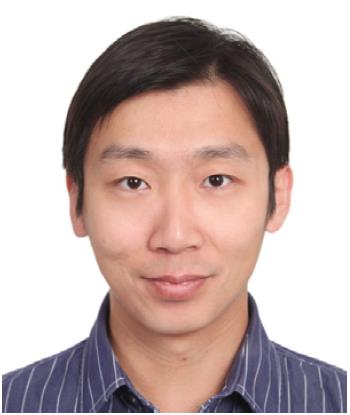}}]{Bo Gao}
  received the Ph.D. degree in computer engineering from Virginia Tech, Blacksburg, VA, USA, in 2014. He was an Assistant Professor with the Institute of Computing Technology, Chinese Academy of Sciences, Beijing, China, from 2014 to 2017. He was a Visiting Researcher with the School of Computing and Communications, Lancaster University, Lancaster, UK, from 2018 to 2019. He is cur-rently an Associate Professor with the School of Computer and Information Technology, Beijing Jiaotong University, Beijing, China. His research interests include wireless net-working, dynamic spectrum sharing, mobile edge computing and multi-agent systems.
\end{IEEEbiography}

\end{document}